\documentclass[preprint,12pt]{elsarticle}

\usepackage{amssymb}
\usepackage{epsfig}
\usepackage{epsf}
\usepackage{booktabs}
\usepackage{amssymb}
\usepackage{multirow}
\usepackage{mathrsfs}
\usepackage{longtable}
\usepackage{lscape}
\usepackage{tabularx}
\usepackage{subfigure}
\newdefinition{definition}{Definition}
\newdefinition{example}{Example}
\newdefinition{remark}{Remark}
\biboptions{square,sort&compress}

\journal{xx}

\begin{document}

\begin{frontmatter}

%% Title, authors and addresses

%% use the tnoteref command within \title for footnotes;
%% use the tnotetext command for the associated footnote;
%% use the fnref command within \author or \address for footnotes;
%% use the fntext command for the associated footnote;
%% use the corref command within \author for corresponding author footnotes;
%% use the cortext command for the associated footnote;
%% use the ead command for the email address,
%% and the form \ead[url] for the home page:
%%
%% \title{Title\tnoteref{label1}}
%% \tnotetext[label1]{}
%% \author{Name\corref{cor1}\fnref{label2}}
%% \ead{email address}
%% \ead[url]{home page}
%% \fntext[label2]{}
%% \cortext[cor1]{}
%% \address{Address\fnref{label3}}
%% \fntext[label3]{}

\title{D numbers theory based game-theoretic framework in adversarial decision making under fuzzy environment}

\author[NWPU]{Xinyang Deng\corref{COR}}
\ead{xinyang.deng@nwpu.edu.cn}
\author[NWPU]{Wen Jiang\corref{COR}}
\ead{jiangwen@nwpu.edu.cn}

\cortext[COR]{Corresponding author.}

\address[NWPU]{School of Electronics and Information, Northwestern Polytechnical University, Xi'an 710072, China}

\begin{abstract}
Adversarial decision making is a particular type of decision making problem where the gain a decision maker obtains as a result of his decisions is affected by the actions taken by others. Representation of alternatives' evaluations and methods to find the optimal alternative are two important aspects in the adversarial decision making. The aim of this study is to develop a general framework for solving the adversarial decision making issue under uncertain environment. By combining fuzzy set theory, game theory and D numbers theory (DNT), a DNT based game-theoretic framework for adversarial decision making under fuzzy environment is presented. Within the proposed framework or model, fuzzy set theory is used to model the uncertain evaluations of decision makers to alternatives, the non-exclusiveness among fuzzy evaluations are taken into consideration by using DNT, and the conflict of interests among decision makers is considered in a two-person non-constant sum game theory perspective. An illustrative application is given to demonstrate the effectiveness of the proposed model. This work, on one hand, has developed an effective framework for adversarial decision making under fuzzy environment; One the other hand, it has further improved the basis of DNT as a generalization of Dempster-Shafer theory for uncertainty reasoning.
\end{abstract}

\begin{keyword}
Adversarial decision making \sep D numbers theory \sep Dempster-Shafer theory \sep Fuzzy set theory \sep Two-person non-constant sum game \sep Uncertainty
%% MSC codes here, in the form: \MSC code \sep code
%% or \MSC[2008] code \sep code (2000 is the default)
\end{keyword}
\end{frontmatter}

\section{Introduction}
In the real world, many scientific and engineering issues can be seen as or converted to decision making problems. Generally, decision making is a complicated process that aims to select an alternative which gives decision makers the highest interests among a variety of options based on the performances or evaluations of alternatives on various criteria including benefit criteria and cost criteria \cite{tzeng2011multiple2011,Bouyssou2013,XDWJIJIS21929}. So far, many technologies and theories have been developed to help decision makers to achieve the optimal decision making, for example analytic hierarchy process (AHP) \cite{Saaty1980,Emrouznejad20175522}, analytic network process (ANP) \cite{saaty2001analyticRWS}, technique for order preference by similarity to ideal solution (TOPSIS) \cite{hwang1981multipleSpringer,Zavadskas1503}. Two important aspects are worthy of particular concern in almost every decision making process. One is the representation of evaluations to alternatives, the other is the method to find the optimal alternative. These two aspects are given special attention in this paper.

Regarding the representation of alternatives' evaluations or performances, essentially it is the issue of information modeling, and there have already many existing means \cite{TverskyA54,HerreraF84337,HerreraF86}. One of the simplest and most straightforward ways is using real numbers to express the performances of alternatives, which is founded on the classical expected utility theory \cite{rabin2013risk241,levy2016expected} and has the most solid theoretical basis. But that means is deficient in specially representing uncertain evaluations, where the uncertainty does not only include randomness but also involve vagueness, imprecision, ambiguity, and so on \cite{kruse2012uncertainty,mendel2017sources245}. With the advance of uncertainty reasoning technologies, many theories have developed to tackle uncertain information and widely used in decision making field, such as theory of interval numbers \cite{moore2009introduction}, possibility theory \cite{dubois2012possibility}, soft set theory \cite{maji2003soft45}, rough set theory \cite{pawlak2012rough}, fuzzy set theory \cite{zadeh1965fuzzy,zimmermann2011fuzzy}, etc. Among them, fuzzy set theory, aiming to cope with the uncertainty of fuzziness, has received much attention and is increasingly flourishing since its intuitive physical meaning and profound philosophical foundation \cite{chan2007global354,tang2017analysis387,liu2013fuzzy621,reza2013fuzzy59,zhang2017novelfuzzy}. On the basis of fuzzy set theory, many branches, such as intuitionistic fuzzy set \cite{Atanassov2012Book} and hesitant fuzzy set \cite{torra2010hesitant256}, constantly emerge, which provide diverse tools to express complicated evaluations in decision making. However, when human beings are involved in the decision making process, overcomplicated data structures will make it hard to get people's authentic evaluations about the alternatives. In balancing the effectiveness and conciseness, fuzzy set theory is a feasible and preferable solution to model the uncertainty in decision making. Therefore, in this paper the decision making under fuzzy environment is of concern.

With respect to the methods of selecting the best alternative in decision making, there are much many theories for various type of decision making problems such as multi-criteria decision making (MCDM), group decision making (GDM), and so on. Among these theories of decision making, many of them consider the situation that the interests of decision makers are basically consistent, but do not pay enough attention on the conflict of decision makers' interests. In many situations the gain a decision maker obtains is not only dependent on his decisions, but also affected by the actions taken by other decision makers \cite{aplakfuzzy2013,xiong2014ambiguous70,deng2014evidential244}. What's more important, these decision makers are conflict of interests. This type of decision making issue is generically referred to as adversarial decision making \cite{yager2008ICCCDadversarial}. In the adversarial decision making, the alternatives of decision makers are changed to strategies that they can choose to achieve their benefits, and a decision maker aims to determine optimal strategy against adversarial opponents. Some researches have been done for the adversarial decision making. For example, Yager \cite{yager2008knowledge231} proposed a knowledge-based approach to adversarial decision making which mainly studies the use of a decision maker's expertise, knowledge, and perceptions, about his adversary to construct a knowledge base about the action his believes his adversary will take. In \cite{pelta2009conflict179}, Pelta and Yager presented a mathematical framework to investigate the balance between inducing confusion and attaining payoff in adversarial decision making. Recent other work, to name but a few, can be found in references \cite{rege2014criminological114,villacorta2011ant,villacorta2012theoretical186,razuri2013adversarial163,froeb2016adversarial593}.

In the most basic form of adversarial decision making, it involves two competitive participants and each of both sides chooses an strategy without knowing the choice of the other. In a perspective of game theory, it constitutes a two-person zero sum or non-constant sum game \cite{washburn2013two201}. Therefore, game theory is feasibly used in researching the adversarial decision making problem \cite{aplakfuzzy2013,xiong2014ambiguous70,yager1999game82}. Based on two aspects of considerations, the representation of human being's uncertain evaluations and methods for decision making, in this paper we suggest a D numbers theory (DNT) based game-theoretic framework for two persons' adversarial decision making under fuzzy environment, where DNT \cite{deng2012DJICS99,xydeng2017DNCR} is a new uncertainty reasoning theory which generalizes Dempster-Shafer theory \cite{Dempster1967,Shafer1976} and already has some applications \cite{fan2016hybrid44,liu2014failure4110,xiao2016intelligent3713518,wang2016route6651,deng2014environmental412}. Within the presented framework, the uncertainty involved in the evaluations to decision makers' strategies is expressed by fuzzy linguistic variables, and the conflict relationship between participants is modelled by a two-person non-constant sum game. Especially, DNT is used to handle and integrate the uncertain evaluations, which fully considers the non-exclusiveness among evaluations expressed fuzzy linguistic variables that is ignored by many previous studies. An illustrative application is given to show the effectiveness of the proposed framework.

The contribution of this paper is two folds. At first, we further improve the basis of DNT by presenting a uniform combination rule for D numbers with complete information or incomplete information, and defining the belief measure and plausibility measure for D numbers, and further clarifying the concept of non-exclusiveness in DNT, and providing a new form for the definition of D numbers. At second, by combining fuzzy set theory, game theory and DNT, we present a new game-theoretic framework for adversarial decision making under fuzzy environment, where the the non-exclusiveness among fuzzy evaluations are fully taken into consideration by exploiting DNT. The rest of this paper is organized as follows. Section \ref{SectPreliminaries} gives a brief introduction about fuzzy set theory, Dempster-Shafer theory, and two-person non-constant sum game. In Section \ref{SectDNT}, the DNT is presented as a generalization of Dempster-Shafer theory on a set with non-exclusive elements. Section \ref{SectProposedFramework} proposes a DNT based game-theoretic framework for adversarial decision making under fuzzy environment, and an illustrative application is given in Section \ref{SectApplication} to show the effectiveness of the proposed framework. Finally, conclusions are given in Section \ref{SectConclusion}.

\section{Preliminaries}\label{SectPreliminaries}

\subsection{Fuzzy set theory}
Fuzzy set theory was first introduced by Zadeh \cite{zadeh1965fuzzy} in 1965 to deal with the uncertainty information. In some real application environments, the states are subjective concepts which are too complex or too ill-defined to be reasonably described in conventional quantitative expressions. In those situation, fuzzy set theory provides an efficiently simple way to express the vagueness or imprecise information \cite{dabbaghian2014sustainability54,jiang2018intuitionistic,song2017uncertainty464,zheng2018evaluation}.

\begin{definition}
Let \( U \) be the universe of discourse, a fuzzy set $\widetilde{A}$ is characterized by a membership function \( \mu _{\widetilde{A}} \)~satisfying
\begin{equation}\label{Eqfuzzyset}
\mu _{\widetilde{A}}: U \to [0,1]
\end{equation}
where $\mu _{\widetilde{A}}(x)$ is called the membership degree of $x \in U$ belonging to fuzzy set $\widetilde{A}$.
\end{definition}

For a finite set \( A = \left\{{\left. {x_1 , \ldots ,x_i , \ldots ,x_n } \right\}} \right. \), the fuzzy set \( (\widetilde{A},\mu _{\widetilde{A}} ) \) is often
denoted by \( \left\{ {\left. {{\raise0.7ex\hbox{${\mu_{\widetilde{A}} (x_1 )}$} \!\mathord{\left/ {\vphantom {{\mu _{\widetilde{A}} (x_1 )} {x_1 }}}\right.\kern-\nulldelimiterspace}\!\lower0.7ex\hbox{${x_1 }$}}, \dots, {\raise0.7ex\hbox{${\mu_{\widetilde{A}} (x_i )}$} \!\mathord{\left/ {\vphantom {{\mu _{\widetilde{A}} (x_i )} {x_i }}}\right.\kern-\nulldelimiterspace}\!\lower0.7ex\hbox{${x_i }$}}, \dots, {\raise0.7ex\hbox{${\mu_{\widetilde{A}} (x_n )}$} \!\mathord{\left/ {\vphantom {{\mu _{\widetilde{A}} (x_n )} {x_n }}}\right.\kern-\nulldelimiterspace}\!\lower0.7ex\hbox{${x_n }$}}} \right\}} \right. \). It is easily found that a fuzzy set is described entirely by its membership function. When \( \mu _{\widetilde{A}}\) takes value from \( \left\{ {\left. {0,1} \right\}} \right. \), fuzzy set \( {\widetilde{A}} \) degenerates into a classical set.

\begin{figure}[htbp]
\begin{center}
\psfig{file=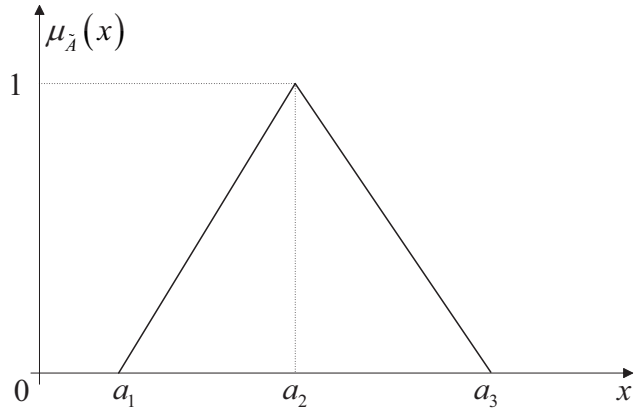,scale=0.80}
\caption{Graphically presentation of the triangular fuzzy number}\label{FigTFN}
\end{center}
\end{figure}

A fuzzy number $\widetilde{A}$ is a fuzzy subset of the real number $R$, and its membership function is
\begin{equation}
\mu _{\widetilde{A}} (x): R \to [0,1]
\end{equation}
where $x$ is a real number and there definitely exists an element $x _0$ such that $\mu _{\widetilde{A}} (x_0) = 1$. Triangular fuzzy numbers are the most widely used fuzzy numbers. A triangular fuzzy number is usually denoted as $\tilde A=(a_1, a_2, a_3)$, as graphically shown in Figure \ref{FigTFN}, which has the following membership function
\begin{equation}
{\mu _{\tilde A}(x)} = \left\{ {\begin{array}{*{20}{c}}
   0, & {x < {a_1}}  \\
   {\frac{{x - {a_1}}}{{{a_2} - {a_1}}}} & {{a_1} \le x \le {a_2}}  \\
   {\frac{{{a_3} - x}}{{{a_3} - {a_2}}}}, & {{a_2} \le x \le {a_3}}  \\
   0, & {x > {a_3}}  \\
\end{array}} \right.
\end{equation}
where $a_1 < a_2 < a_3$.

In theory and practice, fuzzy numbers are usually associated with linguistic variables to express the fuzzy evaluation to objects. A linguistic variable is a variable whose values are represented by words or sentences in a natural or artificial language, for example ``Very Low'', ``Low'', ``Medium'', ``High'', ``Very High'', where there values are usually expressed by fuzzy numbers.

In some applications, it may need to transform a fuzzy number to a real number. There are many approaches proposed for this task. A representative approach is the graded mean integration representation method developed Chou \cite{Chou2003The4510}. Based on that method, given a triangular fuzzy number $\tilde A=(a_1, a_2, a_3)$, its graded mean integration representation is defined as
\begin{equation}\label{EqChouFuzzyIntegration}
P(\tilde A) = {{(a_1  + 4a_2  + a_3 )} \mathord{\left/
 {\vphantom {{(a_1  + 4a_2  + a_3 )} 6}} \right.
 \kern-\nulldelimiterspace} 6}.
\end{equation}

\subsection{Dempster-Shafer theory}
Dempster-Shafer theory (DST) \cite{Dempster1967,Shafer1976}, also called belief function theory or evidence theory, is a popular tool for uncertainty reasoning because of its advantages in expressing uncertainty \cite{AnImprovedAPIN2017,jirouvsek2017new,jiang2017uncertainty} and handling decision making \cite{yager2015dempster80,han2016evaluation461}, clustering and classification \cite{denoeux2016evidential,xu2017data116}, failure mode and effects analysis \cite{gong2018research112,jiang2017failure57}, and other issues \cite{Fontani84593,Elmore476,zheng2018dependence} with uncertainty. This theory needs weaker conditions than the Bayesian theory of probability, so it is often regarded as an extension of the Bayesian theory. As a theory of reasoning under the uncertain environment, DST has an advantage of directly expressing the ``uncertainty" by assigning the basic probability to a set composed of multiple objects, rather than to each of the individual objects. In DST, the information from each information source is seen as a piece of evidence. When there are more than two pieces of evidence, a combination rule is provided to fuse them. For completeness of the explanation, a few basic concepts in DST are introduced as follows.

Let $\Omega$ be a set of $N$ mutually exclusive and collectively exhaustive events, indicated by
\begin{equation}
\Omega  = \{ \theta_1 ,\theta_2 , \cdots ,\theta_i , \cdots ,\theta_N \}
\end{equation}
where set $\Omega$ is called a frame of discernment (FOD). The power set of $\Omega$ is indicated by $2^\Omega$, namely
\begin{equation}
2^\Omega   = \{ \emptyset ,\{ \theta_1 \} , \cdots ,\{ \theta_N \} ,\{ \theta_1
,\theta_2 \} , \cdots ,\{ \theta_1 ,\theta_2 , \cdots ,\theta_i \} , \cdots ,\Omega \}.
\end{equation}
The elements of $2^\Omega$ or subsets of $\Omega$ are called propositions.

\begin{definition}
Let a FOD be $\Omega = \{ \theta_1 ,\theta_2 , \cdots, \theta_N \}$, a mass function defined on $\Omega$ is a mapping $m$ from  $2^\Omega$ to $[0,1]$, formally defined by:
\begin{equation}
m: \quad 2^\Omega \to [0,1]
\end{equation}
which satisfies the following condition:
\begin{eqnarray}
m(\emptyset ) = 0 \quad {\rm{and}} \quad \sum\limits_{A \subseteq \Omega }{m(A) = 1}.
\end{eqnarray}
\end{definition}
In DST, a mass function is also called a basic probability assignment (BPA). The assigned basic probability $m(A)$ measures the belief exactly assigned to $A$ and represents how strongly the evidence supports $A$. If $m(A) > 0$, $A$ is called a focal element, and the union of all focal elements is called the core of the mass function.

Given a BPA, its associated belief measure $Bel$ and plausibility measure $Pl$ express the lower bound and upper bound of the support degree to each proposition in that BPA, respectively. They are defined as
\begin{equation}
Bel(A) = \sum\limits_{B \subseteq A} {m(B)},
\end{equation}
\begin{equation}
Pl(A) = 1 - Bel(\bar A) = \sum\limits_{B \cap A \ne \emptyset }{m(B)},
\end{equation}
where $\bar A = \Omega  - A$. Obviously, $Pl(A) \ge Bel(A)$ for each $A \subseteq \Omega$, and $[Bel(A), Pl(A)]$ is called the belief interval of $A$.

If there are two pieces of evidence, a Dempster's rule of combination is usually used to combine them. This rule assumes that the BPAs are mutually independent.

\begin{definition}
Let $m_1$ and $m_2$ be two BPAs defined on FOD $\Omega$, the Dempster's rule to combine them, denoted by $m = m_1 \oplus m_2$, is defined as follows:
\begin{equation}
m(A) = \left\{ {\begin{array}{*{20}l}
   {\frac{1}{{1 - K}}\sum\limits_{B \cap C = A} {m_1 (B)m_2 (C)} \;,} & {A \ne \emptyset ;}  \\
   {0\;,} & {A = \emptyset }.  \\
\end{array}} \right.
\end{equation}
with
\begin{equation}
K = \sum\limits_{B \cap C = \emptyset } {m_1 (B)m_2 (C)}
\end{equation}
where $K$ is a normalization constant, called conflict coefficient between $m_1$ and $m_2$. Note that the Dempster's rule is only applicable to such two BPAs which satisfy the condition $K < 1$.
\end{definition}

The Dempster's rule plays a very important role in DST, and satisfies commutative and associative properties, i.e., (i) $m_1 \oplus m_2 = m_2 \oplus m_1$ and (ii) $(m_1 \oplus m_2) \oplus m_3 = m_1 \oplus (m_2 \oplus m_3)$. Thus if there exist multiple BPAs, the combination of them can be carried out in a pairwise way with any order.

In order to make decision in terms of a BPA, an approach, called pignistic probability transformation (PPT), is proposed by Smets and Kennes \cite{smets1994transferable662} to derive a distribution of probabilities from the BPA. The PPT function is defined as follows.
\begin{definition}\label{DefPPTFunction}
Let $m$ be a BPA on FOD $\Omega$, a PPT function $BetP_m :\Omega  \to [0,1]$ associated to $m$ is defined by
\begin{equation}
{{BetP}}_m (x) = \sum\limits_{x \in A,A \subseteq \Omega} {\frac{1}{{|A|}}} \frac{{m(A)}}{{1 - m(\emptyset )}}\;,
\end{equation}
where $m(\emptyset ) \ne 1$ and $|A|$ is the cardinality of proposition $A$.
\end{definition}

\subsection{Two-person non-constant sum game}
Game theory provides a mathematical framework to explain and address the interactive decision situations where the aims, goals and preferences of the participating agents are potentially in conflict and it is extensively applied from economics to biology, and to other disciplines \cite{Colman2013,Coupled151,liu2017evidentialgame,WangSAe1601444,CSF103177}. A strategic game consists of three components which are a finite set of players, a nonempty set of strategies for each player, and payoff function for each player in every strategy combination, respectively. The most commonly used solution concept in game theory is that of Nash equilibrium which captures a steady state of the play of a strategic game in which each player holds the correct expectation about the other players' behavior and acts rationally \cite{Osborne1994}.

Two-person non-constant sum game is a kind of widely addressed game, such as iterated prisoner's dilemma game. In this game there are two players, player 1 and player 2. Assuming player 1 has a finite strategy set $S_1$ including $p$ strategies, and layer 2 has a finite strategy set $S_2$ including $q$ strategies. The payoffs of player 1 and player 2 are determined by functions $u_1(s_1, s_2)$ and $u_2(s_1, s_2)$, respectively, where $s_1 \in S_1$ and $s_2 \in S_2$. A combination of players' strategies, denoted as $(s_1^{*}, s_2^{*})$, is a Nash equilibrium of this two-person non-constant sum game if
\begin{equation}\label{NEcondition1}
u_1 (s_1^* ,s_2^* ) \ge u_1 (s_1 ,s_2^* ),\quad \forall s_1  \in S_1
\end{equation}
\begin{equation}\label{NEcondition2}
u_2 (s_1^* ,s_2^* ) \ge u_2 (s_1^* ,s_2 ),\quad \forall s_2  \in S_2
\end{equation}

The Nash equilibrium of a two-person non-constant sum game is a self-enforcing combination of strategies in which no player can gain more by unilaterally deviating from it.

\section{D numbers theory: A generalization of DST on a set with non-exclusive elements}\label{SectDNT}
Although DST has many merits in dealing with uncertain information involving imprecision and ignorance, but it is limited by some hypotheses and constraints that are often hardly satisfied. Related discussions have been detailed in some previous studies \cite{deng2012DJICS99,xydeng2017DNCR,deng2014environmental412}. In summary, there two mainly two aspects. At first, in DST the FOD denoted as $\Omega$ must be composed by mutually exclusive elements. Formally, for any $\theta_i, \theta_j \in \Omega$ they have to meet $\theta_i  \cap \theta_j  = \emptyset$. It is called FOD's exclusiveness hypothesis. By following the hypothesis, if DST is used in fuzzy linguistic environment, any pair of linguistic variables such as ``Good" and ``Very Good" must be exclusive strictly, which however is questionable. Secondly, in DST the sum of basic probabilities or beliefs $m(\cdot)$ in a BPA must be 1, which is called BPA's completeness constraint. However, in some cases, due to lack of knowledge and information, it is possible to obtain an incomplete BPA whose sum of basic probabilities is less than 1. This incompleteness may be caused by an incomplete FOD which corresponds to the case of open world environment, or may be from the lack of information in a closed world environment.

D numbers theory (DNT) is proposed to overcome these limitations mentioned above in DST. This theory is a developing theory in which some key issues still remain unsolved. In \cite{deng2012DJICS99}, Deng first given the definition of D numbers and a combination rule for a special form of D numbers, but that rule is not universal. Besides, in contrast to DST, there are not belief measure and plausibility measure for D numbers so far, so that the lower bound and upper bound of the support in a D number to each proposition can not be derived. In order to let DNT really be a generalization of DST, in this paper a general rule for combining D numbers is proposed, and the belief measure and plausibility measure for D numbers are also developed.

\subsection{Definition of D numbers}

At first, the definition of D numbers is given as follows.
\begin{definition}\label{DefDNumbers}
Let $\Theta$ be a nonempty finite set $\Theta  = \{ F_1 ,F_2 , \cdots ,F_N \}$, a D number is a mapping formulated by
\begin{equation}
D: 2^{\Theta} \to [0,1]
\end{equation}
with
\begin{eqnarray}
\sum\limits_{B \subseteq \Theta } {D(B) \le 1}  \quad {\rm {and}} \quad
D(\emptyset ) = 0
\end{eqnarray}
where $\emptyset$ is the empty set and $B$ is a subset of $\Theta$.
\end{definition}

From the above definition, a D number is defined on a set with non-exclusive elements, which means that any pair of elements in $\Theta$, for example $F_i, F_j \in \Theta$, are not required to be strictly exclusive, i.e. $F_i  \cap F_j  \ne \emptyset$. Here, we still call $\Theta$ as a FOD, but should note that a FOD in DNT is a set consisting of non-exclusive elements. Besides, according to Definition \ref{DefDNumbers}, in a D number the information is not required to be complete. If $\sum\limits_{B \subseteq \Theta } {D(B) = 1}$, we say that the D number is information-complete. By contrast, if $\sum\limits_{B \subseteq \Theta } {D(B) < 1}$ the D number is information-incomplete. The degree of information's completeness in a D number $D$ can be simply expressed by its $Q$ value $Q(D) = \sum\limits_{B \subseteq \Theta } {D(B)}$. In previous studies, we find that a D number with incomplete information is hard to handle mathematically. Facing that, from the view of math a new nonempty set ${ X}$ can be imported to transform a D number with incomplete information to the information-complete case by letting $D ({ X}) = 1 - \sum\limits_{B \subseteq \Theta } {D(B)}$. As a result, a new definition about D numbers is obtained below.
\begin{definition}\label{DefDNumbersEquivalent}
A D number defined on a nonempty finite set $\Theta  = \{ F_1 ,F_2 , \cdots ,F_N \}$ is a mapping $D: 2^{\Theta} \to [0,1]$ satisfying
\begin{eqnarray}
\sum\limits_{B \subseteq \Theta } {D(B) \le 1}  \quad {\rm {and}} \quad
D(\emptyset ) = 0
\end{eqnarray}
and
\begin{equation}
D ({ X}) = 1 - \sum\limits_{B \subseteq \Theta } {D(B)}
\end{equation}
where $\emptyset$ is the empty set, $B$ is a subset of $\Theta$, and $X$ is a nonempty set.
\end{definition}

If $D ({ X}) = 0$ and $\Theta$ becomes a set of mutually exclusive elements, the D number will be completely reduced to a BPA in DST. Therefore, D number is a generalization of BPA. In addition, since in DNT a D number allows $\sum\limits_{B \subseteq \Theta } {D(B) < 1}$, it is very similar with the open world assumption in the transferable belief model (TBM) of DST \cite{Smets1994}. But DNT is essentially different from TBM. At first, in contrast to TBM which implements the open world assumption by letting $m(\emptyset ) > 0$, DNT holds $D(\emptyset ) = 0$ and lets $D ({ X}) > 0$ instead. Secondly, in TBM $\emptyset$ represents all elements that are not included in the FOD, thus the exclusiveness is still hold between FOD and $\emptyset$. However, in DNT $X$ can have an intersection with FOD $\Theta$, even becoming the subset of $\Theta$. Only if $X \not\subset \Theta$ and $D ({ X}) > 0$, the open world assumption is hold. Hence, a D number with incomplete information (i.e. $D ({ X}) > 0$) does not absolutely correspond to the open world environment.

\subsection{FOD's non-exclusiveness in DNT}
As mentioned above, in DNT the elements of FOD $\Theta$ are not required to be mutually exclusive, which means that $F_i$ may be not completely exclusive to $F_j$ for any $F_i, F_j \in \Theta$ and $F_i \cap F_j = \emptyset$. As a natural generalization, the concept of non-exclusiveness can be extended to the subsets of $\Theta$ from the elements of $\Theta$: for two nonempty sets $B_i , B_j \subseteq \Theta$ and $B_i \cap B_j = \emptyset$, $B_i$ may be also not completely exclusive to $B_j$. Further, the non-exclusiveness can be applied to $\Theta  \cup X$ where $D ({ X})$ expresses the incomplete information in DNT as shown in Definition \ref{DefDNumbersEquivalent}. In order to quantitatively represent the non-exclusiveness in $\Theta  \cup X$, a membership function is developed to measure the non-exclusive degrees.
\begin{definition}\label{DefNonExclusiveness}
Given $B_i ,B_j  \in 2^{\Theta  \cup X}$, the non-exclusive degree between $B_i$ and $B_j$ is characterized by a mapping $u_{\neg E}$:
\begin{equation}
u_{\neg E} :2^{\Theta  \cup X}   \times 2^{\Theta  \cup X}   \to [0,1]
\end{equation}
with
\begin{equation}
u_{\neg E} (B_i ,B_j ) = \left\{ \begin{array}{l}
 1,\quad B_i  \cap B_j  \ne \emptyset   \\
 p,\quad B_i  \cap B_j  = \emptyset   \\
 \end{array} \right.
\end{equation}
and
\begin{equation}
u_{\neg E} (B_i ,B_j ) = u_{\neg E} (B_j ,B_i )
\end{equation}
where $0 \le p \le 1$. If letting the exclusive degree between $B_i$ and $B_j$ be denoted as $u_{E}$, then $u_{E} = 1 - u_{\neg E}$.
\end{definition}

According to Definition \ref{DefNonExclusiveness}, the non-exclusive degree between $B_i$ and $B_j$ is 1 if $B_i$ and $B_j$ have intersections, otherwise $u_{\neg E} (B_i ,B_j )$ is $p$ taking a value from $[0,1]$. Obviously, if $u_{\neg E} (B_i ,B_j ) = 0$ for any $B_i  \cap B_j  = \emptyset$, the FOD $\Theta$ in DNT is degenerated to classical FOD in DST. As expressed above, the non-exclusiveness of FOD is one of the most important properties in DNT. In our previous studies \cite{xydeng2017DNCR,Deng2017Fuzzy2086}, a simple approach is developped to calculate all non-exclusive degrees in power set space $2^{\Theta  \cup X}$ if we have the non-exclusive degrees of any pair of elements in ${\Theta  \cup X}$, which is presented as follows
\begin{equation}\label{EqCalNonExclDegeesPowerSet}
u_{\neg E} (B_i ,B_j ) = \mathop {\max }\limits_{x \in B_i ,y \in B_j } \{ u_{\neg E} (x,y)\}
\end{equation}
where $B_i ,B_j  \in 2^{\Theta  \cup X}$. A numerical example is given below to illustrate the approach shown in Eq. (\ref{EqCalNonExclDegeesPowerSet}).

\begin{example}\label{ExampleCalNonExclDegees}
Supposing there is a set of linguistic variables $\Theta  = \{ VP,P,MP,M,MG,G,VG\}$ in which every linguistic variable is represented by a triangular fuzzy number given in Table \ref{TabFuzzyLingTerms} and graphically presented as Figure \ref{FigLinguisticVariables}. The set $\Theta$ is seen as a FOD, and let $D(X)$ represent the possible incomplete information in DNT. For simplicity, it is assumed that $u_{\neg E} (F,X) = 0$ for any $F \in \Theta$. Now the non-exclusive degrees between elements in $2^{\Theta  \cup X}$.

\begin{table}[htbp]
 \begin{center}
    \caption{Fuzzy linguistic variables and corresponding fuzzy numbers}\label{TabFuzzyLingTerms}
    \begin{tabular}{lcc}
    \toprule
    Linguistic variable  & &  Fuzzy number \\
    \midrule
    Very Poor (VP)  & & (0.00, 0.00, 0.25) \\
    Poor (P)  & & (0.10, 0.25, 0.39) \\
    Medium Poor (MP)  & & (0.25, 0.39, 0.53) \\
    Medium (M)  & & (0.39, 0.53, 0.68) \\
    Medium Good (MG)  & & (0.53, 0.68, 0.86) \\
    Good (G)  & & (0.68, 0.86, 0.97) \\
    Very Good (VG)  & & (0.86, 1.00, 1.00) \\
    \bottomrule
    \end{tabular}
 \end{center}
\end{table}

\begin{figure}[htbp]
\begin{center}
\psfig{file=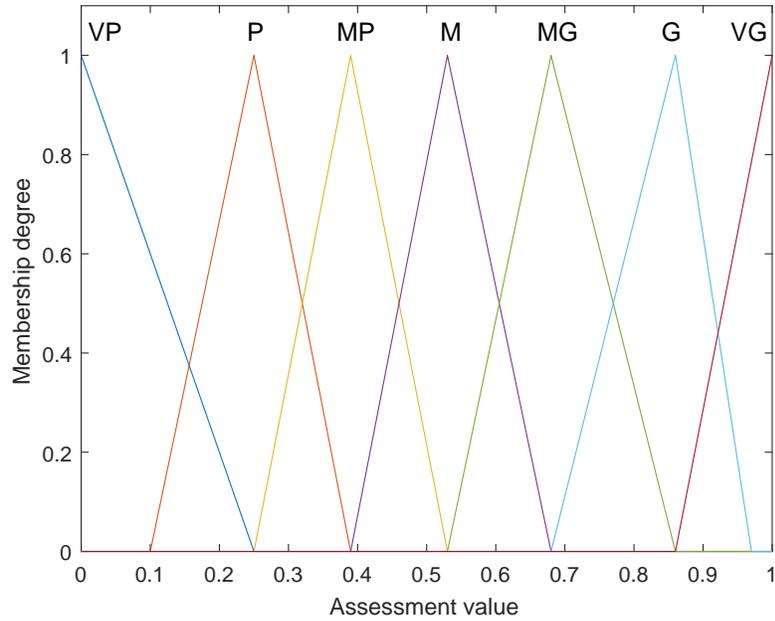,scale=0.80}
\caption{Graphically presentation of fuzzy linguistic variables}\label{FigLinguisticVariables}
\end{center}
\end{figure}

At first, we calculate the non-exclusive degrees between elements in FOD $\Theta$. For that purpose, in this paper the areas of fuzzy numbers are utilized. Let $\widetilde{A}$, $\widetilde{B}$ be two fuzzy numbers with membership functions ${\mu _{\tilde A}(x)}$ and ${\mu _{\tilde B}(x)}$, respectively. And the intersection of $\widetilde{A}$ and $\widetilde{B}$ is denoted as $\tilde A \cap \tilde B$ with membership function $\mu _{\tilde A \cap \tilde B} (x) = \min \{ \mu _{\tilde A} (x),\mu _{\tilde B} (x)\}$. Correspondingly, ${\tilde A \cup \tilde B}$ indicates the union of $\widetilde{A}$ and $\widetilde{B}$ whose membership function is $\mu _{\tilde A \cup \tilde B} (x) = \max \{ \mu _{\tilde A} (x),\mu _{\tilde B} (x)\}$. Then, we define the non-exclusive degree between $\widetilde{A}$ and $\widetilde{B}$ as
\begin{equation}\label{EqNonExclCals}
u_{\neg E} (\widetilde{A} ,\widetilde{B} ) = \frac{{Area}_{{\widetilde{A}} \cap {\widetilde{B}}}}{{ {Area}_{{\widetilde{A}} \cup {\widetilde{B}}}}}
\end{equation}
where ${{Area}_{{\widetilde{A}} \cap {\widetilde{B}}}}$ and ${{ {Area}_{{\widetilde{A}} \cup {\widetilde{B}}}}}$ represent the areas of fuzzy numbers $\tilde A \cap \tilde B$ and $\tilde A \cup \tilde B$, respectively. Based on Eq. (\ref{EqNonExclCals}), and Figure \ref{FigLinguisticVariables}, and the assumed $u_{\neg E} (F,X) = 0, \; \forall F \in \Theta$, each non-exclusive degree between elements in $\Theta \cup X$, therefore, can be obtained as shown in the following matrix
\[\begin{array}{*{20}{c}}
   {} & {\begin{array}{*{20}{c}}
   {\quad VP} & \quad P & {\quad MP} & \quad M & { \quad MG} & \quad G & {\quad VG} & {\quad X} \\
\end{array}}  \\
   {\begin{array}{*{20}{c}}
   {VP}  \\
   P  \\
   {MP}  \\
   M  \\
   {MG}  \\
   G  \\
   {VG}  \\
   X \\
\end{array}} & {\left( {\begin{array}{*{20}{c}}
   1 & {0.116} & 0 & 0 & 0 & 0 & 0 & 0  \\
   {0.116} & 1 & {0.140} & 0 & 0 & 0 & 0 & 0  \\
   0 & {0.140} & 1 & {0.140} & 0 & 0 & 0 & 0  \\
   0 & 0 & {0.140} & 1 & {0.138} & 0 & 0 & 0  \\
   0 & 0 & 0 & {0.138} & 1 & {0.170} & 0 & 0  \\
   0 & 0 & 0 & 0 & {0.170} & 1 & {0.127} & 0  \\
   0 & 0 & 0 & 0 & 0 & {0.127} & 1 & 0  \\
   0 & 0 & 0 & 0 & 0 & 0 & 0 & 1  \\
\end{array}} \right)}  \\
\end{array}\]

At second, once having the above non-exclusive degree matrix of between elements in $\Theta \cup X$ , according to Eq. (\ref{EqCalNonExclDegeesPowerSet}) we can calculate the non-exclusive degree of any pair of elements in $2^{\Theta \cup X}$. For example, as for $\{ VP,P\}$ and $\{ MP,X\}$, we have
\[
\begin{array}{l}
 u_{\neg E} (\{ VP,P\} ,\{ MP,X\} ) \\
  \quad = \max \{ u_{\neg E} (\{ VP,MP\} ),u_{\neg E} (\{ VP,X\} ),u_{\neg E} (\{ P,MP\} ),u_{\neg E} (\{ P,X\} )\}  \\
  \quad = \max \{ 0,\;0,\;0.140,\;0\}  \\
  \quad = 0.140. \\
 \end{array}
\]
The other non-exclusive degrees between elements in $2^{\Theta \cup X}$ can also be derived by the means.
\end{example}

\subsection{Belief measure and plausibility measure for D numbers}
In previous studies, the belief and plausibility for D numbers are not addressed. By considering the non-exclusiveness of FOD in DNT, in this paper we propose a belief measure and a plausibility measure for D numbers as follows.
\begin{definition}\label{}
Let $D$ represent a D number defined on $\Theta$ and $D ({ X})$ express the incomplete information in $D$, for any proposition $A \subseteq \Theta \cup X$, its belief measure $Bel: 2^{\Theta  \cup X}   \to [0,1]$ is defined as
\begin{equation}
Bel(A) = \sum\limits_{B \subseteq A} {D(B)},
\end{equation}
and its plausibility measure $Pl: 2^{\Theta  \cup X}  \to [0,1]$ is defined as
\begin{equation}
Pl(A) = \sum\limits_{B \cap A \ne \emptyset } {D(B)}  + \sum\limits_{B \cap A = \emptyset } {u_{\neg E} (B,A)D(B)},
\end{equation}
where $B \subseteq \Theta \cup X$.
\end{definition}

For the above definition, because $u_{\neg E} (B ,A ) = 1$ for $B  \cap A  \ne \emptyset$, the plausibility measure $Pl$ can be rewritten as
\begin{equation}
Pl(A) = \sum\limits_{B \subseteq \Theta  \cup X} {u_{\neg E} (B,A)D(B)}.
\end{equation}
As same as DST, $[Bel(A), Pl(A)]$ is called the belief interval of $A$ in DNT, which expresses the lower bound and upper bound of support degree to proposition $A$. And it is easy to find that the $Bel$ and $Pl$ for D numbers will degenerate to classical belief measure and plausibility measure in DST if the associated D number is a BPA in fact.

\subsection{Combination rule for D numbers}
How to combine pieces of information is a crucial issue in the theory of information fusion. Ideally, a combination rule for D numbers should be degenerated to the Dempster's rule of combination under a certain conditions, since DNT is designed as a generalization of DST. The rule given in \cite{deng2012DJICS99} is obviously not satisfactory, and its application is limited since that rule is not universal. In a very recent study \cite{xydeng2017DNCR}, we given new combination rules for D numbers with complete information and incomplete information, respectively, from a perspective of conflict redistribution \cite{Dempster1967,yager1987dempster412,dubois1988representation43}. By reexamining the rules given in \cite{xydeng2017DNCR}, we find that they do not well address the case of information-incompleteness and import a D numbers' $Q$ values related function which is difficulty determined in practice. In this paper, by inheriting the idea in \cite{xydeng2017DNCR}, we present a uniform combination rule for D numbers with complete information and incomplete information. The proposed uniform rule is universal for any forms of D numbers and can be totally reduced to Dempster's rule of DST.

\begin{definition}\label{DefDNCR}
Let $D_1$, $D_2$ be two D numbers defined on $\Theta$, and $D_1 ({ X})$, $D_2 ({ X})$ express the incomplete information in $D_1$ and $D_2$ respectively. The combination of $D_1$ and $D_2$, indicated by $D = D_1 \odot  D_2$, is defined by
\begin{equation}\label{EqDefDNCRDA}
D(A) = \left\{ \begin{array}{l}
 0,\quad A = \emptyset  \\
 \frac{1}{{1 - K_D }}\left( \begin{array}{l}
 {\sum\limits_{B \cap C = A} {D_1 (B)D_2 (C)}} \;\; +  \\
 \sum\limits_{\scriptstyle B \cup C = A \hfill \atop
  \scriptstyle B \cap C = \emptyset  \hfill} {u_{\neg E} (B,C)D_1 (B)D_2 (C)}  \\
 \end{array} \right),\quad A \ne \emptyset  \\
 \end{array} \right.
\end{equation}
with
\begin{equation}\label{EqDefDNCRKD}
K_D  = \sum\limits_{B \cap C = \emptyset } {\left( {1 - u_{\neg E} (B,C)} \right)D_1 (B)D_2 (C)},
\end{equation}
where $A, B, C \subseteq {\Theta \cup X}$.
\end{definition}

From Definition \ref{DefDNCR}, the presented rule essentially is a redistribution of exclusive conflict, therefore it is called the exclusive conflict's redistribution (ECR) rule. The proposed ECR rule simultaneously considers the non-exclusiveness of FOD and possible information-incompleteness in D numbers. In this rule, the conflict and incomplete information are represented by $\emptyset$ and $X$, respectively. And it can be completely degenerated to classical Dempster's rule of DST if $u_{\neg E} (B ,C ) = 0$ for any $B  \cap C  = \emptyset$ and $D_1 ({ X}) = 0$, $D_2 ({ X}) = 0$. A numerical example is given below to show the combination process of D numbers by means of the ECR rule.

\begin{example}
Assume there are two D numbers defined on $\Theta = \{a,b\}$:

$D_1(\{a\}) = 0.5$, $D_1(\{b\}) = 0.2$, $D_1(\{a,b\}) = 0.1$, $D_1(\{X\}) = 0.2$;

$D_2(\{a\}) = 0.4$, $D_2(\{b\}) = 0.3$, $D_2(\{a,b\}) = 0.2$, $D_2(\{X\}) = 0.1$.

And the non-exclusive degrees between subsets of ${\Theta \cup X}$ are assumed in the following matrix:
\[
\begin{array}{*{20}c}
   {} & {\begin{array}{*{20}c}
   {\{ X\} } & {\{ b\} } & {\{ b,X\} } & {\{ a\} } & {\{ a,X\} } & {\{ a,b\} } & {\{ a,b,X\} }  \\
\end{array}}  \\
   {\begin{array}{*{20}c}
   {\{ X\} }  \\
   {\{ b\} }  \\
   {\{ b,X\} }  \\
   {\{ a\} }  \\
   {\{ a,X\} }  \\
   {\{ a,b\} }  \\
   {\{ a,b,X\} }  \\
\end{array}} & {\left( {\begin{array}{*{20}c}
   {\rm{1}} & {{\rm{0}}{\rm{.1}}} & {\rm{1}} & {{\rm{0}}{\rm{.2}}} & {\rm{1}} & {{\rm{0}}{\rm{.2}}} & {\rm{1}}  \\
   {{\rm{0}}{\rm{.1}}} & {\rm{1}} & {\rm{1}} & {{\rm{0}}{\rm{.1}}} & {{\rm{0}}{\rm{.1}}} & {\rm{1}} & {\rm{1}}  \\
   {\rm{1}} & {\rm{1}} & {\rm{1}} & {{\rm{0}}{\rm{.2}}} & {\rm{1}} & {\rm{1}} & {\rm{1}}  \\
   {{\rm{0}}{\rm{.2}}} & {{\rm{0}}{\rm{.1}}} & {{\rm{0}}{\rm{.2}}} & {\rm{1}} & {\rm{1}} & {\rm{1}} & {\rm{1}}  \\
   {\rm{1}} & {{\rm{0}}{\rm{.1}}} & {\rm{1}} & {\rm{1}} & {\rm{1}} & {\rm{1}} & {\rm{1}}  \\
   {{\rm{0}}{\rm{.2}}} & {\rm{1}} & {\rm{1}} & {\rm{1}} & {\rm{1}} & {\rm{1}} & {\rm{1}}  \\
   {\rm{1}} & {\rm{1}} & {\rm{1}} & {\rm{1}} & {\rm{1}} & {\rm{1}} & {\rm{1}}  \\
\end{array}} \right)}  \\
\end{array}
\]

The combination result of $D_1$ and $D_2$ can be obtained through the following steps. At first, an intersection/union table is calculated as shown Table \ref{ExOneValidDNTIntersection}, where each item is derived by either ${D_1 (B)D_2 (C)}$ assigned to ${B \cap C}$ if ${B \cap C \ne \emptyset}$ or ${u_{\neg E} (B,C)D_1 (B)D_2 (C)}$ assigned to ${B \cup C}$ if ${B \cap C = \emptyset}$.

\begin{table}[htbp]
    \begin{center}
    \caption{Intersection/union table in combining $D_1$ and $D_2$}\label{ExOneValidDNTIntersection}
    \begin{tabular}{l|rrrr}
    \toprule
    $D_1 \odot D_2$ &  $D_2(\{a\}) = 0.4$     &    $D_2(\{b\}) = 0.3$  &  $D_2(\{a,b\}) = 0.2$   &    $D_2(\{X\}) = 0.1$  \\
    \midrule
    $D_1(\{a\}) = 0.5$  &　$\{a\}$ (0.2)   &　$\{a,b\}$ (0.015)  &　$\{a\}$ (0.1)  &　$\{a,X\}$ (0.01) \\
    $D_1(\{b\}) = 0.2$  &　$\{a,b\}$ (0.008)  &　$\{b\}$ (0.06)  &　$\{b\}$ (0.04)  &　$\{b,X\}$ (0.002) \\
    $D_1(\{a,b\}) = 0.1$  &　$\{a\}$ (0.04)  &　$\{b\}$ (0.03)  &　$\{a,b\}$ (0.02)  &　$\{a,b,X\}$ (0.002) \\
    $D_1(\{X\}) = 0.2$  &　$\{a,X\}$ (0.016)  &　$\{b,X\}$ (0.006)  &　$\{a,b,X\}$ (0.008)  &　$\{X\}$ (0.02) \\
    \bottomrule
    \end{tabular}
    \end{center}
\end{table}

Then according to Eq. (\ref{EqDefDNCRKD}), we can have the conflict coefficient $K_D  = 0.423$. Therefore, in terms of Eq. (\ref{EqDefDNCRDA}) the final result of combining $D_1$ and $D_2$ is obtained
\[
\begin{array}{l}
 D(\{ a\} ) = 0.589, \\
 D(\{ b\} ) = 0.225, \\
 D(\{ a,b\} ) = 0.075, \\
 D(\{ X\} ) = 0.035, \\
 D(\{ a,X\} ) = 0.045, \\
 D(\{ b,X\} ) = 0.014, \\
 D(\{ a,b,X\} ) = 0.017. \\
 \end{array}
\]
\end{example}

It must be pointed out that the proposed ECR rule in Definition \ref{DefDNCR} satisfies the commutative property, i.e. ${D_1} \odot {D_2} = {D_2} \odot {D_1}$, but does not preserve the associative property, namely $({D_1} \odot {D_2}) \odot {D_3} \ne {D_1} \odot ({D_2} \odot {D_3}) \ne ({D_1} \odot {D_3}) \odot {D_2}$. In the theory of information fusion, there are two main schemes when combining pieces of information \cite{florea2009robust102}. One is the aggregating scheme where evidences represent different opinions about the same event, the other is the updating scheme in which evidences express sequential opinions about a dynamic event. For the commutative and associative rules, such as the Dempster's rule, there is no difference between these two schemes. But for the non-associative rules, the two schemes provide different results. In DNT, the ECR rule is naturally suitable for the updating scheme, but is not appropriate to be used in the aggregating scheme. Facing that, a weighted average combination (WAC) method is suggested to combine multiple D numbers about the same event based on the proposed ECR rule.

Suppose there are $n$ D numbers indicated by $D_1$, $D_2$, $\cdots$, $D_n$, and every D number is given a weighting factor $w_i$, $i=1,\cdots, n$, satisfying $\sum\limits_{i = 1}^n {{w_i}}  = 1$. At first, the averaging D number among $D_1$, $D_2$, $\cdots$, $D_n$ is defined as
\begin{equation}\label{EqWACAverage}
\bar D(B) = \sum\limits_{i = 1}^n {{w_i}{D_i}(B)}.
\end{equation}
Then, the result of combining $D_1$, $D_2$, $\cdots$, $D_n$ is obtained by
\begin{equation}\label{EqWACCombine}
D_{WAC} =  \mathop  \odot \limits_{i = 1}^n {{\bar D}_i}
\end{equation}
where $\bar D_i = \bar D $, $i=1, \cdots, n$, and $\odot$ is the ECR rule given in Definition \ref{DefDNCR}.

\section{Proposed D numbers theory based game-theoretic framework}\label{SectProposedFramework}
\begin{figure}[htbp]
\begin{center}
\psfig{file=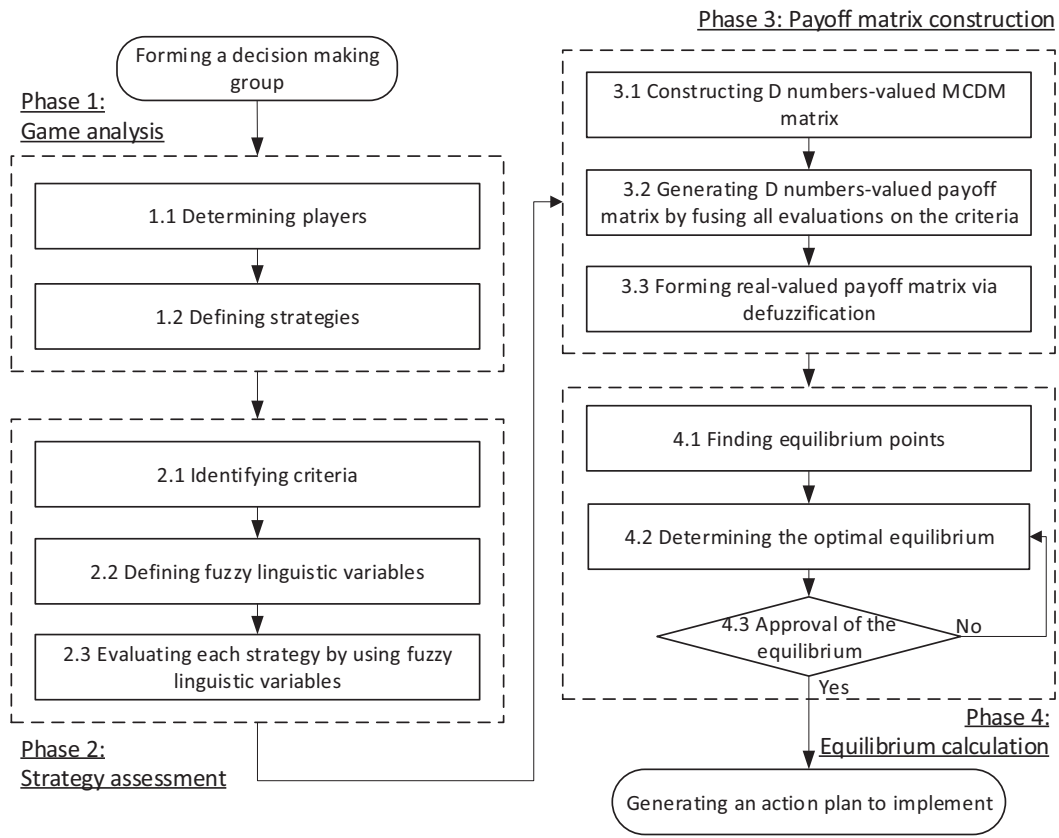,scale=0.85}
\caption{Flow diagram of the proposed DNT based game-theoretic framework}\label{FigFramework}
\end{center}
\end{figure}

In this section, a DNT based game-theoretic framework is proposed for adversarial decision making under uncertain environment, whose flow diagram is graphically shown in Figure \ref{FigFramework}. Underlying this framework, a two-person non-constant sum game is considered. The framework mainly consists of four phases including ``Game analysis", ``Strategy assessment", ``Payoff matrix construction", and ``Equilibrium calculation", respectively. In the following text, we briefly describe these phases, leaving the details in the next section to explain via an illustrative application.

\begin{enumerate}[{Phase} 1.]
  \item By means of a formed decision making group (DMG), in the phase of game analysis the players in the game are easily determined at first. Then, the DMG will define a strategy set for each player by analyzing the current decision making situation.
  \item In the phase of strategy assessment, the main task is to evaluate the strategies of each player. At first, the decision making criteria are identified as the basis of assessment. Secondly, by considering an uncertain decision making environment, we define fuzzy linguistic variables to express the uncertainty in the evaluations. Based on the criteria and linguistic variables determined previously, each strategy for each player is evaluated by the DMG. As a result, a multi-experts and multi-criteria decision making matrix with fuzzy evaluations is formed for every strategy of each player.
  \item The third phase is to construct the payoff matrix of the game. For the decision making matrix of each strategy, firstly we collect the evaluations from multi-experts on every criterion and express them in a D number, which is to construct a D numbers-valued MCDM matrix for every strategy. Then, we combine these D numbers on different criteria for every strategy to generate a D numbers-valued payoff matrix, where the fusion of evaluations on multi-criteria is implemented. At last, in order to calculate the equilibria of the game, the generated D numbers-valued payoff matrix is converted to a real-valued payoff matrix via the defuzzification of D numbers.
  \item In the phase of equilibrium calculation, all equilibria of the game will be found at first according to the real-valued payoff matrix obtained above. Evidently, there are two cases about the found equilibria. One is that there is only an equilibrium point, in this case the unique equilibrium naturally become the outcome of the game. The other is that there are at least two equilibrium points, in the case we must choose an optimal equilibrium point as the outcome of the game. In game theory, different approach may yield different optimal equilibrium point, where each optimal equilibrium is associated with different point of view on ``optimal". The determination of the optimal equilibrium is on the basis of practical applications and specific demands, therefore it is not the focus of the study. By some means an optimal equilibrium is assumed to be determined, the DMG will re-examine the equilibrium  prudently. If the optimal equilibrium is not satisfactory, a new equilibrium point will be required; Otherwise, if it is approved, the optimal equilibrium will be used to generate the action plan for the adversarial decision making.
\end{enumerate}

\section{An illustrative application}\label{SectApplication}
In this section, an example adapted from reference \cite{aplakfuzzy2013} is given to illustrate the process of applying the proposed DNT based game-theoretic framework and verify its effectiveness. For more details about the application, please refer to \cite{aplakfuzzy2013}.

\begin{figure}[htbp]
\begin{center}
\psfig{file=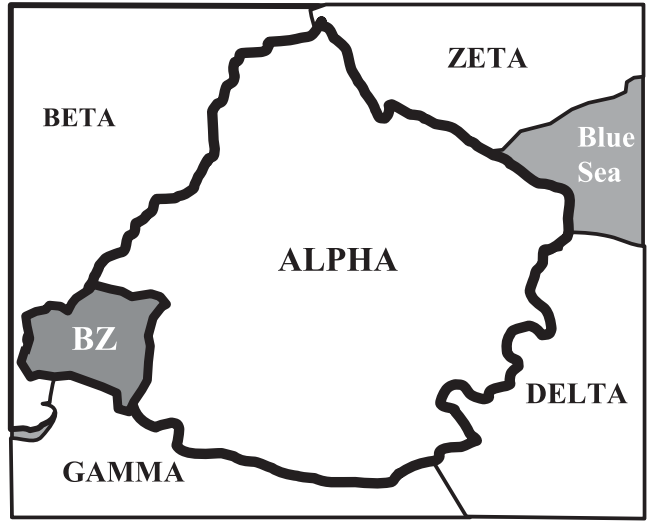,scale=1.4}
\caption{An example of adversarial decision making about territorial dispute \cite{aplakfuzzy2013}}\label{FigApplication}
\end{center}
\end{figure}

In this application, it is assumed that there is a territorial dispute between two countries Alpha and Beta about the belonging of a buffer zone (BZ), as shown in Figure \ref{FigApplication}. Recently, Beta unilaterally claims that the BZ belongs to it. For this reason, Alpha call for an international aid against Beta's demand. United Nations send and located international peace support force to control the area. Regarding to this situation, the conflict between countries is formalized in two-person non-constant game theory and analyzed by the proposed DNT based game-theoretic framework. Before the analysis, it is assumed a DMG consisting of ten decision makers (DM) is established in advance.

\subsection{Game analysis}
In this phase, the DMG is to determine the players in the game and define possible strategies for each player. Regarding the players, obviously there are two players, namely Alpha and Beta. Regarding the strategies, by following reference \cite{aplakfuzzy2013} it is assumed that there are five strategies for Alpha and four strategies for Beta.

The strategies of Alpha are
\begin{itemize}
  \item ${AS}_1$: Main and rear area control
  \item ${AS}_2$: Sector control
  \item ${AS}_3$: Area control as whole
  \item ${AS}_4$: Strong area control as whole
  \item ${AS}_5$: Area control with local forces
\end{itemize}

The strategies of Beta are
\begin{itemize}
  \item ${BS}_1$: Attack by itself
  \item ${BS}_2$: Corporate attack
  \item ${BS}_3$: Passive attitude
  \item ${BS}_4$: Ownership of conflict zone (BZ)
\end{itemize}

\subsection{Strategy assessment}
In the phase of strategy assessment, at first we identify key criteria so as to evaluate the strategies for players. In previous study \cite{aplakfuzzy2013}, six key criteria have been identified, including ``Management", ``Protection", ``Mobility", ``Logistic", ``Flexibility", ``Simplicity", and a fuzzy weigh for every criterion has been also determined, as shown in Table \ref{TabCriteriaWeights}.

\begin{table}[htbp]
 \begin{center}
  \caption{Fuzzy weights for criteria}\label{TabCriteriaWeights}
    \begin{tabular}{lcc}
    \toprule
    Criterion  & &  Fuzzy weight \\
    \midrule
    Management ($C_1$) & & (0.53, 0.91, 1.00) \\
    Protection ($C_2$) & & (0.39, 0.66, 1.00) \\
    Mobility ($C_3$) & & (0.10, 0.41, 0.68) \\
    Logistic ($C_4$) & & (0.25, 0.60, 1.00) \\
    Flexibility ($C_5$) & & (0.39, 0.82, 1.00) \\
    Simplicity ($C_6$) & & (0.10, 0.49, 1.00) \\
    \bottomrule
    \end{tabular}
    \end{center}
\end{table}

Secondly, a group of fuzzy linguistic terms should be defined for the sake of expressing DM's uncertain evaluations to the strategies. In the paper, the linguistic variables given in Table \ref{TabFuzzyLingTerms} are employed directly. These linguistic variables, ranked from ``Very Poor" (VP) to ``Very Good" (VG), constitute a scale of seven grades, as graphically shown in Figure \ref{FigLinguisticVariables}.

Thirdly, according to the criteria identified above, the DMG is to evaluate these strategies for each player by using the defined fuzzy linguistic variables. Different from conventional MCDM process, each evaluation matrix is established by given opposite player's strategies from a game theory perspective since the application is in an adversarial decision making environment. For example, Table \ref{TabASEvaluationscondiseringB1} gives the evaluations to Alpha's all strategies in the case of Beta choosing ${BS}_1$ as its strategy, which is abbreviated as ${BS}_1$ Case for short. And since there are ten decision makers in the DMG, Table \ref{TabASEvaluationscondiseringB1} is also an evaluation matrix of a group decision making. By means of this way, we can obtain the other evaluation matrices, namely ${BS}_2$ Case, ${BS}_3$ Case, ${BS}_4$ Case for Alpha, and ${AS}_1$ Case, ${AS}_2$ Case, ${AS}_3$ Case, ${AS}_4$ Case, ${AS}_5$ Case for Beta. In this paper, all of the evaluation matrices are from literature \cite{aplakfuzzy2013}.

\begin{table}[htbp]\footnotesize
 \begin{center}
  \caption{AS's evaluations according to criteria with linguistic variables (${BS}_1$ Case)}\label{TabASEvaluationscondiseringB1}
    \begin{tabular}{cccccccccccc}
    \toprule
    Criterion & Strategy & DM 1 & DM 2 & DM 3 & DM 4 & DM 5 & DM 6 & DM 7 & DM 8 & DM 9 & DM 10 \\
    \midrule
    $C_1$ & ${AS}_1$ & MG & MG & M & MG & G & G & G & MG & G & G \\
    {} & ${AS}_2$ & G & G & MG & M & VG & G & MG & MG & VG & M \\
    {} & ${AS}_3$ & P & VP & VP & VP & VP & VP & VP & P & VP & VP \\
    {} & ${AS}_4$ & VP & P & P & VP & MP & P & P & P & P & MP \\
    {} & ${AS}_5$ & P & VP & VP & P & VP & VP & VP & P & P & VP \\
    $C_2$ & ${AS}_1$ & M & MP & M & M & MP & P & G & MG & MG & MG \\
    {} & ${AS}_2$ & G & M & M & G & M & M & MG & MP & M & MP \\
    {} & ${AS}_3$ & MP & VP & VP & P & VP & P & VP & P & MP & P \\
    {} & ${AS}_4$ & MP & MP & VP & P & MP & M & MP & P & M & VP \\
    {} & ${AS}_5$ & VP & MP & MP & VP & MP & VP & VP & VP & VP & MP \\
    $C_3$ & ${AS}_1$ & VG & VG & MG & VG & VG & G & VG & MG & MG & VG \\
    {} & ${AS}_2$ & VG & MG & G & VG & MG & VG & VG & VG & MG & MG \\
    {} & ${AS}_3$ & MP & M & P & P & M & P & P & P & P & P \\
    {} & ${AS}_4$ & MG & MG & M & MG & MG & MG & M & M & MP & G \\
    {} & ${AS}_5$ & MP & VP & P & P & VP & M & VP & VP & P & M \\
    $C_4$ & ${AS}_1$ & VG & VG & VG & MG & VG & VG & VG & VG & G & VG \\
    {} & ${AS}_2$ & G & MG & G & MG & M & M & MG & G & G & P \\
    {} & ${AS}_3$ & G & MP & MP & P & M & P & MP & MP & M & MP \\
    {} & ${AS}_4$ & M & M & M & MP & P & MP & M & P & M & P \\
    {} & ${AS}_5$ & MP & VP & MP & P & M & P & P & MP & P & MP \\
    $C_5$ & ${AS}_1$ & MP & VG & MP & M & VP & G & M & MG & G & MP \\
    {} & ${AS}_2$ & G & G & G & MG & M & MG & M & M & MG & MG \\
    {} & ${AS}_3$ & MP & MP & VP & P & VP & M & VP & P & P & MP \\
    {} & ${AS}_4$ & MP & M & P & P & P & P & P & MP & VP & P \\
    {} & ${AS}_5$ & VP & MP & P & VP & MP & MG & VP & P & VP & M \\
    $C_6$ & ${AS}_1$ & MG & M & G & MG & G & MG & G & G & MG & G \\
    {} & ${AS}_2$ & MP & MG & M & MP & M & MG & MG & MG & MG & MG \\
    {} & ${AS}_3$ & M & M & MG & M & MG & M & M & MG & M & MG \\
    {} & ${AS}_4$ & MP & MP & M & MG & M & MP & MP & M & M & MP \\
    {} & ${AS}_5$ & P & P & MP & MG & MP & MP & P & MG & M & M \\
    \bottomrule
    \end{tabular}
    \end{center}
\end{table}

\subsection{Payoff matrix construction}
Through the above two phases, nine game-theoretic evaluation matrices have been established for each player's strategies. But it is difficult to find the solution (or equilibrium point) of the adversarial decision making problem directly based on these evaluation matrices. We must construct a payoff matrix for this game in terms of the evaluation matrices. During the process, it will implement the fusion of evaluations from multiple decision makers and multiple criteria, where DNT is used to deal with these non-exclusive fuzzy evaluations.

Firstly, for every strategy we transform the fuzzy evaluations on the same criterion given by ten decision makers into a D number to realize the fusion of multiple experts. For example, in Table \ref{TabASEvaluationscondiseringB1}, for Alpha's strategy ${AS}_1$, on $C_1$ four decision makers give the evaluation of MG, one gives M, and the other five give G. By assuming these decision makers have the same importance, hence a D number is generated as follows
\[
\left\{ \begin{array}{l}
 D(MG) = 0.4, \\
 D(M) = 0.1, \\
 D(G) = 0.5 \\
 \end{array} \right.
\]
Here, the above D number is simply denoted as (\{MG\}, 0.4; \{M\}, 0.1; \{G\}, 0.5). Via this means, all fuzzy evaluations can be integrated and re-expressed by D numbers. In this application, all the obtained D numbers are information-complete. As a result, every fuzzy evaluation matrix, like ${BS}_1$ Case in Table \ref{TabASEvaluationscondiseringB1}, is transformed to a D numbers-valued MCDM matrix. Table \ref{TabDNMCDMMatrixBS1} shows the D numbers-valued MCDM matrix derived from Table \ref{TabASEvaluationscondiseringB1}.

\begin{table}[htbp]\footnotesize
 \begin{center}
  \caption{D numbers-valued MCDM matrix (${BS}_1$ Case)}\label{TabDNMCDMMatrixBS1}
    \begin{tabular}{ccl}
    \toprule
    Criterion & Strategy & Evaluation \\
    \midrule
    $C_1$ & ${AS}_1$ & (\{M\}, 0.1; \{MG\}, 0.4; \{G\}, 0.5) \\
    {} & ${AS}_2$ &   (\{M\}, 0.2; \{MG\}, 0.3; \{G\}, 0.3; \{VG\}, 0.2)\\
    {} & ${AS}_3$ &   (\{VP\}, 0.8; \{P\}, 0.2)\\
    {} & ${AS}_4$ &   (\{VP\}, 0.2; \{P\}, 0.6; \{MP\}, 0.2)\\
    {} & ${AS}_5$ &   (\{VP\}, 0.6; \{P\}, 0.4)\\
    $C_2$ & ${AS}_1$ &   (\{P\}, 0.1; \{MP\}, 0.2; \{M\}, 0.3; \{MG\}, 0.3; \{G\}, 0.1) \\
    {} & ${AS}_2$ &   (\{MP\}, 0.2; \{M\}, 0.5; \{MG\}, 0.1; \{G\}, 0.2)\\
    {} & ${AS}_3$ &   (\{VP\}, 0.4; \{P\}, 0.4; \{MP\}, 0.2) \\
    {} & ${AS}_4$ &   (\{VP\}, 0.2; \{P\}, 0.2; \{MP\}, 0.4; \{M\}, 0.2) \\
    {} & ${AS}_5$ &   (\{VP\}, 0.6; \{MP\}, 0.4) \\
    $C_3$ & ${AS}_1$ &  (\{MG\}, 0.3; \{G\}, 0.1; \{VG\}, 0.6) \\
    {} & ${AS}_2$ &  (\{MG\}, 0.4; \{G\}, 0.1; \{VG\}, 0.5) \\
    {} & ${AS}_3$ &  (\{P\}, 0.7; \{MP\}, 0.1; \{M\}, 0.2) \\
    {} & ${AS}_4$ &  (\{MP\}, 0.1; \{M\}, 0.3; \{MG\}, 0.5; \{G\}, 0.1) \\
    {} & ${AS}_5$ &  (\{VP\}, 0.4; \{P\}, 0.3; \{MP\}, 0.1; \{M\}, 0.2) \\
    $C_4$ & ${AS}_1$ & (\{MG\}, 0.1; \{G\}, 0.1; \{VG\}, 0.8) \\
    {} & ${AS}_2$ &  (\{P\}, 0.1; \{M\}, 0.2; \{MG\}, 0.3; \{G\}, 0.4) \\
    {} & ${AS}_3$ &  (\{P\}, 0.2; \{MP\}, 0.5; \{M\}, 0.2; \{G\}, 0.1) \\
    {} & ${AS}_4$ &  (\{P\}, 0.3; \{MP\}, 0.2; \{M\}, 0.5) \\
    {} & ${AS}_5$ &  (\{VP\}, 0.1; \{P\}, 0.4; \{MP\}, 0.4; \{M\}, 0.1) \\
    $C_5$ & ${AS}_1$ & (\{VP\}, 0.1; \{MP\}, 0.3; \{M\}, 0.2; \{MG\}, 0.1; \{G\}, 0.2; \{VG\}, 0.1) \\
    {} & ${AS}_2$ &  (\{M\}, 0.3; \{MG\}, 0.4; \{G\}, 0.3) \\
    {} & ${AS}_3$ &  (\{VP\}, 0.3; \{P\}, 0.3; \{MP\}, 0.3; \{M\}, 0.1) \\
    {} & ${AS}_4$ &  (\{VP\}, 0.1; \{P\}, 0.6; \{MP\}, 0.2; \{M\}, 0.1) \\
    {} & ${AS}_5$ &  (\{VP\}, 0.4; \{P\}, 0.2; \{MP\}, 0.2; \{M\}, 0.1; \{MG\}, 0.1) \\
    $C_6$ & ${AS}_1$ & (\{M\}, 0.1; \{MG\}, 0.4; \{G\}, 0.5) \\
    {} & ${AS}_2$ &  (\{MP\}, 0.2; \{M\}, 0.2; \{MG\}, 0.6) \\
    {} & ${AS}_3$ &  (\{M\}, 0.6; \{MG\}, 0.4) \\
    {} & ${AS}_4$ &  (\{MP\}, 0.5; \{M\}, 0.4; \{MG\}, 0.1) \\
    {} & ${AS}_5$ &  (\{P\}, 0.3; \{MP\}, 0.3; \{M\}, 0.2; \{MG\}, 0.2) \\
    \bottomrule
    \end{tabular}
    \end{center}
\end{table}

Secondly, the evaluations to every strategy on different criteria, now expressed by D numbers, are fused to derive an integrated evaluation to each strategy. This is a typical fusion process of evaluations on multiple criteria, but the data is expressed by D numbers. Within the process, the ECR based WAC method, shown in Eqs. (\ref{EqWACAverage}) and (\ref{EqWACCombine}), is used to combine multiple D numbers, where the weight factors of D numbers are derived from the fuzzy weights of criteria given in Table \ref{TabCriteriaWeights}. At the firs step, the fuzzy weights given in Table \ref{TabCriteriaWeights} are transformed to crisp values by using Eq. (\ref{EqChouFuzzyIntegration}), we have $P(C_1 ) =  0.862$, $P(C_2 ) =  0.672$, $P(C_3 ) =  0.403$, $P(C_4 ) =  0.608$, $P(C_5 ) =  0.778$, $P(C_6 ) =  0.510$. At the second step, these crisp values are normalized as the weight factors of the criteria, namely
\[
\begin{array}{l}
 w_{C_1 }  = 0.225,\quad w_{C_2 }  = 0.175,\quad w_{C_3 }  = 0.105, \\
 w_{C_4 }  = 0.159,\quad w_{C_5 }  = 0.203,\quad w_{C_6 }  = 0.133. \\
 \end{array}
\]
Now we can fuse the evaluations on all criteria for every strategy. Let's use strategy ${AS}_1$ as the example. From Table \ref{TabDNMCDMMatrixBS1}, given Beta's strategy ${BS}_1$ the evaluations to ${AS}_1$ are
\begin{itemize}
  \item On $C_1$: $D_{AS_1|BS_1 }^{C_1 } (\{ M\} ) = 0.1,D_{AS_1|BS_1 }^{C_1 } (\{ MG\} ) = 0.4,D_{AS_1|BS_1 }^{C_1 } (\{ G\} ) = 0.5$;
  \item On $C_2$: $D_{AS_1|BS_1 }^{C_2 } (\{ P\} ) = 0.1,D_{AS_1|BS_1 }^{C_2 } (\{ MP\} ) = 0.2,D_{AS_1|BS_1 }^{C_2 } (\{ M\} ) = 0.3,D_{AS_1|BS_1 }^{C_2 } (\{ MG\} ) = 0.3,D_{AS_1|BS_1 }^{C_2 } (\{ G\} ) = 0.1$;
  \item On $C_3$: $D_{AS_1|BS_1 }^{C_3 } (\{ MG\} ) = 0.3,D_{AS_1|BS_1 }^{C_3 } (\{ G\} ) = 0.1,D_{AS_1 |BS_1 }^{C_3 } (\{ VG\} ) = 0.6$;
  \item On $C_4$: $D_{AS_1|BS_1 }^{C_4 } (\{ MG\} ) = 0.1,D_{AS_1 |BS_1}^{C_4 } (\{ G\} ) = 0.1,D_{AS_1 |BS_1 }^{C_4 } (\{ VG\} ) = 0.8$;
  \item On $C_5$: $ D_{AS_1|BS_1 }^{C_5 } (\{ VP\} ) = 0.1,D_{AS_1|BS_1 }^{C_5 } (\{ MP\} ) = 0.3,D_{AS_1 |BS_1 }^{C_5 } (\{ M\} ) = 0.2,D_{AS_1|BS_1 }^{C_5 } (\{ MG\} ) = 0.1,D_{AS_1|BS_1 }^{C_5 } (\{ G\} ) = 0.2,D_{AS_1|BS_1 }^{C_5 } (\{ VG\} ) = 0.1$;
  \item On $C_6$: $D_{AS_1|BS_1 }^{C_6 } (\{ M\} ) = 0.1,D_{AS_1|BS_1 }^{C_6 } (\{ MG\} ) = 0.4,D_{AS_1|BS_1 }^{C_6 } (\{ G\} ) = 0.5$.
\end{itemize}
Based on the weight factors calculated above, the averaging D number of $D_{AS_1|BS_1 }^{C_1 }$, $D_{AS_1|BS_1 }^{C_2 }$, $D_{AS_1|BS_1 }^{C_3 }$, $D_{AS_1|BS_1 }^{C_4 }$, $D_{AS_1|BS_1 }^{C_5 }$, $D_{AS_1 |BS_1}^{C_6 }$, denoted as $\bar D_{AS_1|BS_1 }$, therefore, is
\[
\begin{array}{l}
 \bar D_{AS_1|BS_1 } (\{ VG\} ) = 0.210, \\
 \bar D_{AS_1|BS_1 } (\{ G\} ) = 0.263, \\
 \bar D_{AS_1|BS_1 } (\{ MG\} ) = 0.263, \\
 \bar D_{AS_1|BS_1 } (\{ M\} ) = 0.129, \\
 \bar D_{AS_1|BS_1 } (\{ MP\} ) = 0.096, \\
 \bar D_{AS_1|BS_1 } (\{ P\} ) = 0.018, \\
 \bar D_{AS_1|BS_1 } (\{ VP\} ) = 0.020. \\
 \end{array}
\]
Then, based on the non-exclusive degrees calculated in Example \ref{ExampleCalNonExclDegees} and  Eq. (\ref{EqWACCombine}), given Beta's strategy ${BS}_1$  the integrated evaluation to strategy ${AS}_1$, denoted as $D_{AS_1|BS_1 }$, can be obtained
\[
\begin{array}{l}
 D_{AS_1|BS_1 } (\{ VG\} ) = {0.089,} \\
 D_{AS_1|BS_1 } (\{ G\} ) = {0.378,} \\
 D_{AS_1|BS_1 } (\{ G,VG\} ) = {0.037,} \\
 D_{AS_1|BS_1 } (\{ MG\} ) = {0.344,} \\
 D_{AS_1|BS_1 } (\{ MG,G\} ) = {0.089,} \\
 D_{AS_1|BS_1 } (\{ MG,G,VG\} ) = {0.011,} \\
 D_{AS_1|BS_1 } (\{ M\} ) = {0.020,} \\
 D_{AS_1|BS_1 } (\{ M,MG\} ) = {0.019,} \\
 D_{AS_1|BS_1 } (\{ M,MG,G\} ) = 0.007, \\
 D_{AS_1|BS_1 } (\{ M,MG,G,VG\} ) = 0.001, \\
 D_{AS_1|BS_1 } (\{ MP\} ) = 0.002, \\
 D_{AS_1|BS_1 } (\{ MP,M\} ) = {0.001,} \\
 D_{AS_1|BS_1 } (\{ MP,M,MG\} ) = {0.001} \\
 \end{array}
\]
where the terms whose beliefs are smaller than 0.001 are not displayed. From the perspective of game theory, $D_{AS_1 |BS_1}$ represents the payoff of Alpha's strategy ${AS}_1$ given ${BS}_1$ chosen by Beta. By this means, the payoff of each strategy, either Alpha's or Beta's, given opposite player's a strategy, can be obtained. These payoffs are used to form a D numbers-valued payoff matrix for the game, as shown in Table \ref{TabDNPayoffMatrix}.

\begin{landscape}
\begin{table}[htbp]
  \centering
  \caption{D numbers-valued payoff matrix for the game}\label{TabDNPayoffMatrix}
    \begin{tabular}{cccccc}
    \toprule
     {} & {} & \multicolumn{4}{c}{Beta} \\
     \cline{3-6}
     {} & {} & {${BS}_1$} & {${BS}_2$} & {${BS}_3$} & {${BS}_4$}   \\
    \midrule
     {Alpha} & ${AS}_1$ & $(D_{AS_1 |BS_1}, D_{BS_1 |AS_1})$ & $(D_{AS_1 |BS_2}, D_{BS_2 |AS_1})$ & $(D_{AS_1 |BS_3}, D_{BS_3 |AS_1})$ & $(D_{AS_1 |BS_4}, D_{BS_4 |AS_1})$ \\
     {} & ${AS}_2$ & $(D_{AS_2 |BS_1}, D_{BS_1 |AS_2})$ & $(D_{AS_2 |BS_2}, D_{BS_2 |AS_2})$ & $(D_{AS_2 |BS_3}, D_{BS_3 |AS_2})$ & $(D_{AS_2 |BS_4}, D_{BS_4 |AS_2})$ \\
     {} & ${AS}_3$ & $(D_{AS_3 |BS_1}, D_{BS_1 |AS_3})$ & $(D_{AS_3 |BS_2}, D_{BS_2 |AS_3})$ & $(D_{AS_3 |BS_3}, D_{BS_3 |AS_3})$ & $(D_{AS_3 |BS_4}, D_{BS_4 |AS_3})$ \\
     {} & ${AS}_4$ & $(D_{AS_4 |BS_1}, D_{BS_1 |AS_4})$ & $(D_{AS_4 |BS_2}, D_{BS_2 |AS_4})$ & $(D_{AS_4 |BS_3}, D_{BS_3 |AS_4})$ & $(D_{AS_4 |BS_4}, D_{BS_4 |AS_4})$ \\
     {} & ${AS}_5$ & $(D_{AS_5 |BS_1}, D_{BS_1 |AS_5})$ & $(D_{AS_5 |BS_2}, D_{BS_2 |AS_5})$ & $(D_{AS_5 |BS_3}, D_{BS_3 |AS_5})$ & $(D_{AS_5 |BS_4}, D_{BS_4 |AS_5})$ \\
    \bottomrule
    \end{tabular}
\end{table}
\end{landscape}

\begin{figure}[htbp]
\begin{center}
\psfig{file=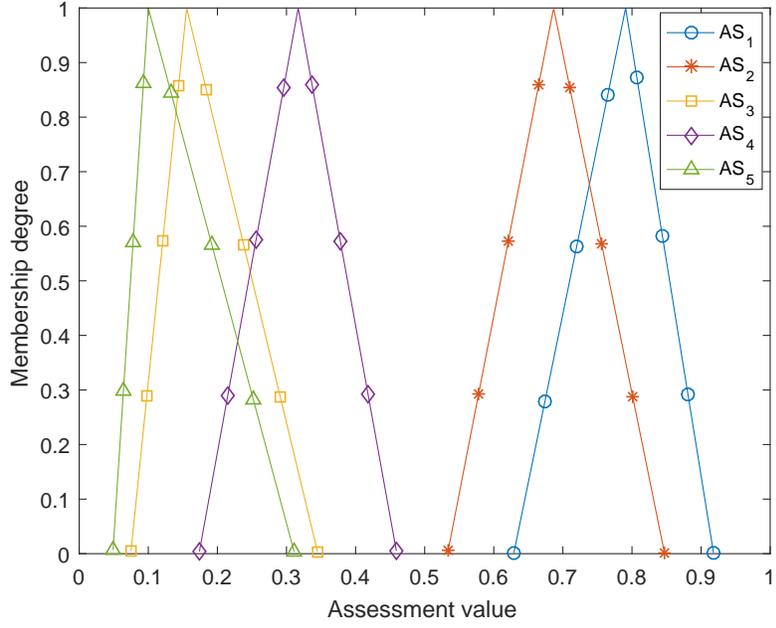,scale=0.80}
\caption{Fuzzy payoffs of Alpha's strategies given ${BS}_1$ chosen by Beta}\label{FigASFuzzyPayoffsGivenBS1}
\end{center}
\end{figure}

Thirdly, we will transform the D numbers-valued payoff matrix to a real-valued payoff matrix since currently it is difficult to find the equilibria of a game with D numbers-valued payoffs. The process is decomposed into three steps. At the first step, we simply use the PPT function in Definition \ref{DefPPTFunction} to derive a distribution of probabilities from each D numbers-valued payoff. For example, the distribution of probabilities derived from $D_{AS_1 |BS_1}$ is
\[
\begin{array}{l}
 DIS_{AS_1 |BS_1 } (VP) = 0.000, \\
 DIS_{AS_1 |BS_1 } (P) = 0.000, \\
 DIS_{AS_1 |BS_1 } (MP) = 0.002, \\
 DIS_{AS_1 |BS_1 } (M) = 0.033, \\
 DIS_{AS_1 |BS_1 } (MG) = 0.405, \\
 DIS_{AS_1 |BS_1 } (G) = 0.448, \\
 DIS_{AS_1 |BS_1 } (VG) = 0.112. \\
 \end{array}
\]
At the second step, every distribution of probabilities ${DIS}_{AS_i |BS_j }$, as a form of payoff, is further transformed to a fuzzy payoff by the following formula
\begin{equation}\label{EqFuzzyPayoffFunction}
\tilde P_{AS_i |BS_j }  = \sum\limits_{A \in \Theta } {DIS_{AS_i |BS_j } (A) \times \mu_A (x)}
\end{equation}
where $\Theta  = \{ VP,P,MP,M,MG,G,VG\}$ whose elements are fuzzy linguistic variables given in Table \ref{TabFuzzyLingTerms}, and $\mu_A (x)$ is the membership function of fuzzy number $A$, $A \in \Theta$. In terms of Eq. (\ref{EqFuzzyPayoffFunction}), as for $DIS_{AS_1 |BS_1 }$ we have
\[
\tilde P_{AS_1 |BS_1 }  = (0.629,0.791,0.918).
\]
Similarly, given ${BS}_1$ the fuzzy payoffs of Alpha's other strategies are calculated
\[
\begin{array}{l}
 \tilde P_{AS_2 |BS_1 }  = (0.533,0.687,0.847), \\
 \tilde P_{AS_3 |BS_1 }  = (0.075,0.156,0.346), \\
 \tilde P_{AS_4 |BS_1 }  = (0.173,0.317,0.460), \\
 \tilde P_{AS_5 |BS_1 }  = (0.049,0.100,0.312). \\
 \end{array}
\]
Figure \ref{FigASFuzzyPayoffsGivenBS1} graphically shows the above fuzzy payoffs of Alpha's strategies given ${BS}_1$ chosen by Beta. At the third step, these fuzzy payoffs will be transformed to real-valued payoffs via the defuzzification. In this paper, the centroid defuzzification method \cite{Yager1981A242} is used, which is presented as follows
\begin{equation}\label{EqDefuzzification}
defuzzy(\tilde P) = \frac{{\int {x\mu _{\tilde P} (x)dx} }}{{\int {\mu _{\tilde P} (x)dx} }}.
\end{equation}
Therefore, given ${BS}_1$ chosen by Beta the real-valued payoffs of Alpha's strategies are
\[
\begin{array}{l}
 defuzzy(\tilde P_{AS_1 |BS_1 } ) = 0.779, \\
 defuzzy(\tilde P_{AS_2 |BS_1 } ) = 0.689, \\
 defuzzy(\tilde P_{AS_3 |BS_1 } ) = 0.192, \\
 defuzzy(\tilde P_{AS_4 |BS_1 } ) = 0.317, \\
 defuzzy(\tilde P_{AS_5 |BS_1 } ) = 0.154. \\
 \end{array}
\]
From the results, if Beta chooses ${BS}_1$ as its strategy, Alpha's best response is strategy ${AS}_1$. Through the above three steps, the D numbers-valued payoff matrix in Table \ref{TabDNPayoffMatrix} is transformed to a real-valued payoff matrix which is shown in Table \ref{TabRealValuedPayoffMatrix}.

\begin{table}[htbp]
  \centering
  \caption{Real-valued payoff matrix for the game}\label{TabRealValuedPayoffMatrix}
    \begin{tabular}{cccccc}
    \toprule
     {} & {} & \multicolumn{4}{c}{Beta} \\
     \cline{3-6}
     {} & {} & {${BS}_1$} & {${BS}_2$} & {${BS}_3$} & {${BS}_4$}   \\
    \midrule
     {Alpha}      & ${AS}_1$ &  (0.779, 0.586) &  (0.700, 0.508) &  (0.742, 0.609) &  (0.680, 0.757)  \\
               {} & ${AS}_2$ &  (0.689, 0.561) &  (0.830, 0.356) &  (0.648, 0.524) &  (0.377, 0.842)  \\
               {} & ${AS}_3$ &  (0.192, 0.676) &  (0.218, 0.531) &  (0.790, 0.835) &  (0.662, 0.828)  \\
               {} & ${AS}_4$ &  (0.317, 0.633) &  (0.347, 0.578) &  (0.734, 0.786) &  (0.499, 0.734)  \\
               {} & ${AS}_5$ &  (0.154, 0.753) &  (0.200, 0.606) &  (0.937, 0.781) &  (0.854, 0.717)  \\
    \bottomrule
    \end{tabular}
\end{table}

\subsection{Equilibrium calculation}
In the phase, the equilibrium points of the game can be easily calculated based on the obtained real-valued payoff matrix. From Table \ref{TabRealValuedPayoffMatrix}, it is found that there is only an equilibrium point $({AS}_5 , {BS}_3 )$. Therefore, it is not necessary to conduct the selection of optimal equilibrium, and $({AS}_5 , {BS}_3 )$ is automatically approved to be the equilibrium to generate every player's action plan. Thus, in the example of territorial dispute between two counties Alpha and Beta, the players will choose ${AS}_5$ and ${BS}_3$ as their action plans, respectively.

\subsection{Comparison and discussion}
Let us compare the result obtained by the proposed DNT based framework or method with that by a TOPSIS based method developed in \cite{aplakfuzzy2013}. On the basis of the same data, a real-valued payoff matrix for the application is also derived by using the TOPSIS based method \cite{aplakfuzzy2013}, which is shown in Table \ref{TabPayoffMatrixTOPSIS}. In terms of Tables \ref{TabRealValuedPayoffMatrix} and \ref{TabPayoffMatrixTOPSIS}, we can compare the DNT based method and TOPSIS based method as follows.

At first, from Table \ref{TabPayoffMatrixTOPSIS} the equilibrium point is also $({AS}_5 , {BS}_3 )$ by using the TOPSIS based method, which is the same as the result of this paper that uses the DNT based method. Both of these two methods have obtained correct result. Therefore, the two methods are both effective for the adversarial decision making under fuzzy environment.

\begin{table}[htbp]
  \centering
  \caption{Obtained real-valued payoff matrix for the application by using a TOPSIS based method in \cite{aplakfuzzy2013}}\label{TabPayoffMatrixTOPSIS}
    \begin{tabular}{cccccc}
    \toprule
     {} & {} & \multicolumn{4}{c}{Beta} \\
     \cline{3-6}
     {} & {} & {${BS}_1$} & {${BS}_2$} & {${BS}_3$} & {${BS}_4$}   \\
    \midrule
     {Alpha}      & ${AS}_1$ &  (0.519, 0.465) &  (0.510, 0.432) &  (0.500, 0.489) &  (0.507, 0.546) \\
               {} & ${AS}_2$ &  (0.510, 0.472) &  (0.554, 0.423) &  (0.489, 0.451) &  (0.451, 0.552) \\
               {} & ${AS}_3$ &  (0.352, 0.477) &  (0.373, 0.431) &  (0.505, 0.488) &  (0.487, 0.540) \\
               {} & ${AS}_4$ &  (0.375, 0.465) &  (0.391, 0.480) &  (0.517, 0.489) &  (0.469, 0.500) \\
               {} & ${AS}_5$ &  (0.338, 0.507) &  (0.322, 0.441) &  (0.554, 0.520) &  (0.531, 0.513) \\
    \bottomrule
    \end{tabular}
\end{table}

Secondly, by looking into the payoffs of players at the equilibrium point $({AS}_5 , {BS}_3 )$, the DNT based method and TOPSIS based method both obtain that Alpha has gained the maximum payoff among its all possible payoffs, while Beta can just get a relatively high payoff at the equilibrium point. Since the goals of players are usually in conflict in a game, players do not simultaneously achieve the maximum interests. Both of the two methods have effectively reflected this characteristic of adversarial decision making.

Thirdly, we examine the ranking of strategies. Table \ref{TabAlphaStraRanking} shows the rankings of Alpha's strategies for every Beta's strategy in the two methods, and Table \ref{TabBetaStraRanking} gives the rankings of Beta's strategies for all Alpha's strategies. In these tables, the differences between rankings obtained by the DNT based method and TOPSIS based method are highlighted. Since the determination of equilibria is only based on the top 1 strategies (as known as best responses in game theory) of Alpha and Beta, we just pay the attention on the best responses. As seen from Tables \ref{TabAlphaStraRanking} and \ref{TabBetaStraRanking}, in most cases, except ${AS}_3$ and ${AS}_4$, the best responses obtained by the two methods for Alpha and Beta are the same. Given ${AS}_3$ or ${AS}_4$ as Alpha's strategy, Beta's best response is ${BS}_3$ in terms of the DNT based method, while ${BS}_4$ by using TOPSIS based method. From Table \ref{TabAlphaStraRanking}, as for player Alpha, strategy ${AS}_5$ has two times to be the best response, which is the most among Alpha's strategies, either using the DNT based method or using TOPSIS based method. While for player Beta, from Table \ref{TabBetaStraRanking} ${BS}_3$ is the strategy having the most times (three times) to be the best response if using the DNT based method, and it is ${BS}_4$ (four times) if using the TOPSIS based method. Hence, the most likely strategies of Alpha and Beta, determined by the TOPSIS based method, are respectively ${AS}_5$ and ${BS}_4$, while the equilibrium found through the same method is $({AS}_5 , {BS}_3 )$. As written in \cite{aplakfuzzy2013}, ``Any information gathered during execution phase, may orient DM (decision maker) to other directions". In using the TOPSIS based method, the found most likely strategies and equilibrium point are inconsistent to top decision maker, which may lead to a deviation from the rational actions in real decision making. In contrast, by using the DNT based method the derived most likely strategies of Alpha and Beta are ${AS}_5$ and ${BS}_3$, respectively, which are completely consistent with the equilibrium $({AS}_5 , {BS}_3 )$. In this sense, the results obtained by the DNT based method are more reasonable than that of the TOPSIS based method.

\begin{landscape}
\begin{table}[htbp]
  \centering
  \caption{The rankings of Alpha's strategies for every Beta's strategy based on different methods}\label{TabAlphaStraRanking}
    \begin{tabular}{ccccccccccccc}
    \toprule
     {} & {} & \multicolumn{11}{c}{Beta} \\
     \cline{3-13}
     {} & {} & \multicolumn{2}{c}{${BS}_1$} & {} & \multicolumn{2}{c}{${BS}_2$} & {} & \multicolumn{2}{c}{${BS}_3$} & {} & \multicolumn{2}{c}{${BS}_4$}   \\
     \cline{3-4}\cline{6-7}\cline{9-10}\cline{12-13}
     {} & {} & DNT & TOPSIS & {} & DNT & TOPSIS & {} & DNT & TOPSIS & {} & DNT & TOPSIS \\
    \midrule
          {Alpha} & ${AS}_1$ & 1 & 1 & {} & 2 & 2 & {} & \underline{\textbf{3}} & \underline{\textbf{4}} & {} & 2 & 2 \\
               {} & ${AS}_2$ & 2 & 2 & {} & 1 & 1 & {} & 5 & 5 & {} & 5 & 5 \\
               {} & ${AS}_3$ & 4 & 4 & {} & 4 & 4 & {} & \underline{\textbf{2}} & \underline{\textbf{3}} & {} & 3 & 3 \\
               {} & ${AS}_4$ & 3 & 3 & {} & 3 & 3 & {} & \underline{\textbf{4}} & \underline{\textbf{2}} & {} & 4 & 4 \\
               {} & ${AS}_5$ & 5 & 5 & {} & 5 & 5 & {} & 1 & 1 & {} & 1 & 1 \\
    \bottomrule
    \end{tabular}
\end{table}
\end{landscape}

\begin{landscape}
\begin{table}[htbp]
  \centering
  \caption{The rankings of Beta's strategies for every Alpha's strategy based on different methods}\label{TabBetaStraRanking}
    \begin{tabular}{cccccccccccccccc}
    \toprule
     {} & {} & \multicolumn{14}{c}{Alpha} \\
     \cline{3-16}
     {} & {} & \multicolumn{2}{c}{${AS}_1$} & {} & \multicolumn{2}{c}{${AS}_2$} & {} & \multicolumn{2}{c}{${AS}_3$} & {} & \multicolumn{2}{c}{${AS}_4$} & {} & \multicolumn{2}{c}{${AS}_5$}  \\
     \cline{3-4}\cline{6-7}\cline{9-10}\cline{12-13}\cline{15-16}
     {} & {} & DNT & TOPSIS & {} & DNT & TOPSIS & {} & DNT & TOPSIS & {} & DNT & TOPSIS& {} & DNT & TOPSIS  \\
    \midrule
           {Beta} & ${BS}_1$ & 3 & 3 & {} & 2 & 2 & {} & 3 & 3 & {} & \underline{\textbf{3}} & \underline{\textbf{4}} & {} & \underline{\textbf{2}} & \underline{\textbf{3}} \\
               {} & ${BS}_2$ & 4 & 4 & {} & 4 & 4 & {} & 4 & 4 & {} & \underline{\textbf{4}} & \underline{\textbf{3}} & {} & 4 & 4 \\
               {} & ${BS}_3$ & 2 & 2 & {} & 3 & 3 & {} & \underline{\textbf{1}} & \underline{\textbf{2}} & {} & \underline{\textbf{1}} & \underline{\textbf{2}} & {} & 1 & 1 \\
               {} & ${BS}_4$ & 1 & 1 & {} & 1 & 1 & {} & \underline{\textbf{2}} & \underline{\textbf{1}} & {} & \underline{\textbf{2}} & \underline{\textbf{1}} & {} & \underline{\textbf{3}} & \underline{\textbf{2}} \\
    \bottomrule
    \end{tabular}
\end{table}
\end{landscape}

\section{Conclusions}\label{SectConclusion}
Decision making is of catholic concern in many fields. In a traditional decision making process, there is only an overall goal among decision makers and the attention is usually given on the selection of optimal one among multiple decision making alternatives. In this study, our focus is on the issue of adversarial decision making where the competition not only takes place in alternatives but also occurs in decision makers. Since there is conflict of interests among decision makers, the aim of adversarial decision making is to determine the best alternative for each decision maker against other adversarial opponents. This paper considers a basic form of adversarial decision making which contains two competitive participants under uncertain environment. We have proposed a new framework for this type of decision making problem by utilizing fuzzy set theory, game theory and DNT. The proposed DNT based game-theoretic framework are made up by four phases including game analysis, strategy assessment, payoff matrix construction, and equilibrium calculation. Within the framework, the uncertainty involved in decision makers' evaluations is handled by using fuzzy set theory, and DNT, as a new uncertainty reasoning theory that generalizes Dempster-Shafer theory and has been further improved in the study, is used to integrate these fuzzy evaluations which are not mutually exclusive, and a two-person non-constant sum game is constructed to model the conflict of interests between decision makers from a game theory perspective. An illustrative application is given to show the effectiveness of the framework. In the future study, the adversarial decision making with multiple competitive participants is worthy to be considered by further extending the proposed framework.

\section*{Acknowledgments}
The authors thank Dr. Hakan Soner Aplak for providing the data. The work is partially supported by National Natural Science Foundation of China (Program Nos. 61703338, 61671384), Natural Science Basic Research Plan in Shaanxi Province of China (Program No. 2016JM6018), Project of Science and Technology Foundation, Fundamental Research Funds for the Central Universities (Program No. 3102017OQD020).

%% The Appendices part is started with the command \appendix;
%% appendix sections are then done as normal sections
%% \appendix

%% \section{}
%% \label{}

%% References
%%
%% Following citation commands can be used in the body text:
%% Usage of \cite is as follows:
%%   \cite{key}         ==>>  [#]
%%   \cite[chap. 2]{key} ==>> [#, chap. 2]
%%

%% References with bibTeX database:

\bibliographystyle{elsarticle-num}
\bibliography{references}

%% Authors are advised to submit their bibtex database files. They are
%% requested to list a bibtex style file in the manuscript if they do
%% not want to use elsarticle-num.bst.

%% References without bibTeX database:

% \begin{thebibliography}{00}

%% \bibitem must have the following form:
%%   \bibitem{key}...
%%

% \bibitem{}

% \end{thebibliography}

\end{document}